\documentclass{ecai}
\usepackage{times}
\usepackage{graphicx}
\usepackage{latexsym}
\usepackage[colorlinks,linkcolor=blue]{hyperref}
\usepackage{color}
\usepackage{amsmath}

\begin{document}
\title{Improving Abstractive Text Summarization with History Aggregation}
\author{Pengcheng Liao$^{1,2}$ , Chuang Zhang$^2$ , Xiaojun Chen$^2$ \and Xiaofei Zhou$^2$}
\maketitle
\nocite{*}
\bibliographystyle{ecai}

\begin{abstract}
 Recent neural sequence to sequence models have provided feasible solutions for abstractive summarization. However, such models are still hard to tackle long text dependency in the summarization task. A high-quality summarization system usually depends on strong encoder which can refine important information from long input texts so that the decoder can generate salient summaries from the encoder’s memory. In this paper, we propose an aggregation mechanism based on the Transformer model to address the challenge of long text representation. Our model can review history information to make encoder hold more memory capacity. Empirically, we apply our aggregation mechanism to the Transformer model and experiment on CNN/DailyMail dataset to achieve higher quality summaries compared to several strong baseline models on the ROUGE metrics.
\end{abstract}
\footnotetext[1]{School of Cyber Security, University of Chinese Academy of Sciences, Beijing, China}
\footnotetext[2]{Institute of Information Engineering, Chinese Academy of Sciences, Beijing, China}
\footnotetext[3]{Corresponding author: Chuang Zhang, \href{zhangchuang@iie.ac.cn}{zhangchuang@iie.ac.cn}}

\section{INTRODUCTION}
\label{sec:introduction}
The task of text summarization is automatically compressing long text to a shorter version while keeping the salient information. It can be divided into two approaches: extractive and abstractive summarization. The extractive approach usually selects sentences or phrases from source text directly. On the contrary, the abstractive approach first understands the semantic information of source text and generates novel words not appeared in source text. Extractive summarization is easier, but abstractive summarization is more like the way humans process text. This paper focuses on the abstractive approach. 
Unlike other sequence generation tasks in NLP(Natural Language Processing) such as NMT(Neural Machine Translation), in which the lengths of input and output text are close, the summarization task exists severe imbalance on the lengths. It means that the summarization task must model long-distance text dependencies.

As RNNs have the ability to tackle time sequence text, variants of sequence to sequence model\cite{sutskever2014sequence}  based on them have emerged on a large scale and can generate promising results. To solve the long-distance text dependencies, Bahdnau et al.\cite{bahdanau2014neural} first propose the attention mechanism which allows each decoder step to refer to all encoder hidden states.
Rush et al.\cite{rush2015a} first incorporate attention mechanism to summarization task. There are also other attention-based models to ease the problem of long input texts for summarization task, like Bahdnau attention\cite{see2017get}, hierarchical attention\cite{nallapati2016abstractive}, graph-based attention\cite{tan2017abstractive} and simple attention\cite{lopyrev2015generating}.
Celikyilmaz et al.\cite{celikyilmaz2018deep} segment text and encode segmented text independently then broadcast their encoding to others. Though these systems are promising, they exhibit undesirable behaviors such as producing inaccurate factual details and repeating themselves as it is hard to decide where to attend and where to ignore for one-pass encoder.\par

Modeling an effective encoder for representing a long text is still a challenge, and we are committed to solving long text dependency problems by aggregation mechanism. The key idea of the aggregation mechanism is to collect history information then computes attention between the encoder final hidden states and history information to re-distribute the encoder final states. It suggests that the encoder can read long input texts a few times to understand the text clearly. We build our model by reconstructing the Transformer model\cite{vaswani2017attention} via incorporating our novel aggregation mechanism. Empirically, we first analyze the features of summarization and translation dataset. Then we experiment with different encoder and decoder layers, and the results reveal that the ability of the encoder layer is more important than the decoder layer, which implies that we should focus more on the encoder. Finally, we experiment on CNN/DailyMail dataset, and our model generates higher quality summaries compared to strong baselines of Pointer Generator and Transformer models on ROUGE metrics and human evaluations. \par 

The main contributions of this paper are as follows:
\begin{itemize}
	\item We put forward a novel aggregation mechanism to redistribute context states of text with collected history information. Then we equip the Transformer model with the aggregation mechanism.
	\item Our model outperforms 1.01 ROUGE-1, 0.30 ROUGE-2 and 1.27 ROUGE-L scores on CNN/DailyMail dataset and 5.31 ROUGE-1, 4.56 ROUGE-2 and 5.19 ROUGE-L scores on our build Chinese news dataset compared to Transformer baseline model.
	
\end{itemize}
 \par

\section{RELATED WORK}
\label{sec:realtedwork}
In this section, we first introduce extractive summarization then introduce abstractive summarization.
 
\subsection{Extractive Summarization}
Extractive summarization aims to select salient sentences from source documents directly. This method is always modeled as a sentence ranking problem via selecting sentences with high scores\cite{kupiec1995a}, sequence labeling(binary label) problem\cite{conroy2001text} or integer linear programmers\cite{woodsend2010automatic}. The models above mostly leverage manual engineered features, but they are now replaced by the neural network to extract features automatically.
Cheng et al.\cite{cheng2016neural} get sentence representation using convolutional neural network(CNN) and document representation using recurrent neural network(RNN) and then select sentences/words using hierarchical extractor.
Nallapati et al.\cite{nallapati2016summarunner:} treat the summarization as a sequence labeling task. They get sentence and document representations using RNNs and after a classification layer, each sentence will get a label which indicates whether this sentence should be selected.
Zhou et al.\cite{zhou2018neural} present a model for extractive summarization by jointly learning score and select sentences.
Zhang et al. \cite{zhang2018neural} put forward a latent variable model to tackle the problem of sentence label bias.
\subsection{Abstractive Summarization}
\label{subsec:abstract summarization}
Abstractive summarization aims to rewrite source texts with understandable semantic meaning. Most methods of this task are based on sequence to sequence models.
Rush et al.\cite{rush2015a} first incorporate the attention mechanism to abstractive summarization and achieve state of the art scores on DUC-2004 and Gigaword datasets.
Chopra et al.\cite{chopra2016abstractive} improve the model performance via RNN decoder.
Nallapati et al.\cite{nallapati2016abstractive} adopt a hierarchical network to process long source text with hierarchical structure.
Gu et al.\cite{gu2016incorporating} are the first to show that a copy mechanism can take advantage of both extractive and abstractive summarization by copying words from the source text (extractive summarization) and generating original words (abstractive summarization).
See et al.\cite{see2017get} incorporate copy and coverage mechanisms to avoid generating inaccurate and repeated words.
Celikyilmaz et al.\cite{celikyilmaz2018deep} split text to paragraph and apply encoder to each paragraph, then broadcast paragraph encoding to others.
Recently, Vaswani et al.\cite{vaswani2017attention} give a new view of sequence to sequence model. It employs the self-attention to replace RNN in sequence to sequence model and uses multi-head attention to capture different semantic information.

Lately, more and more researchers focus on combine abstractive and extractive summarization. 
Hsu et al.\cite{hsu2018a} build a unified model by using inconsistency loss. 
Gehrmann et al.\cite{gehrmann2018bottom} first train content-selector to select and mask salient information then train the abstractive model (Pointer Generator) to generate abstractive summarization.    

\section{MODEL}
\label{sec:model}
\begin{figure*}
	
	\centerline{\includegraphics[width=\linewidth]{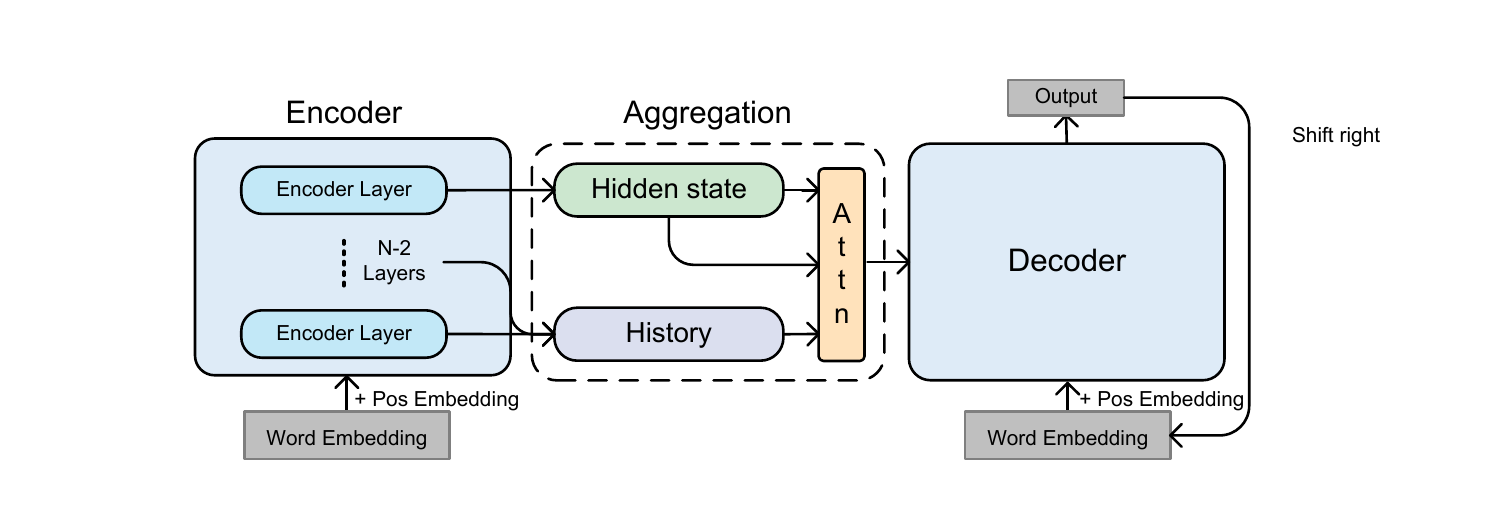}}
	\caption{Aggregation Transformer model overview. Compared with the Transformer baseline model, we apply the aggregation layer between encoder and decoder. The aggregation layer can collect history information to redistribute the encoder's final hidden states.}
	\label{fig:base_model}
\end{figure*}
In this section, we first describe the attention mechanism and the Transformer baseline model, after that, we introduce the pointer and BPE mechanism. Our novel aggregation mechanism is described in the last part. The code for our model is available online.\footnote[1]{\url{https://github.com/Pc-liao/Transformer_agg}}\par 

\textbf{Notation} We have pairs of texts $\{X,Y\}$, where $d \in X$ is a long text and  $y \in Y$ is the summary of corresponding $d$. The lengths of $d$ and $y$ is $ld$ and $ly$ respectively. Each text $d$ is composed by a sequence of words $w$, and we embed word $w$ into vector $e$. So we represent document $d$ with embedding vector $\{e^1, e^2, ..., e^{ld}\}$ and we can get representation of $y$ the same as $d$.

\subsection{Attention Mechanism}
\label{ssec:attention}
The attention mechanism is widely used in text summarization models as it can produce word significance distribution in source text for disparate decode steps.
Bahdanau et al.\cite{bahdanau2014neural} first propose the attention mechanism where attention weight distribution can be calculated:
\begin{equation}
e_i^t=v^\top tanh(w_s s_t+w_hh_i + b_i^t)
\end{equation}
\begin{equation}
Attention^t  =  softmax(e^t)
\end{equation}

Where $h_i$ is the encoder hidden states in $i$th word, $s_t$ is decoder hidden states at time step $t$. Vector $v,w_s,w_h$ and scalar $b_i^t$ are learnable parameters. $Attention^t$ is probability distribution that represents the importance of different source words for decoder at time step $t$.\par
Transformer redefines attention mechanism more concisely.
In practice, we compute the attention function on a set of queries simultaneously, packed together into a matrix $Q$. The keys and values are also packed together into matrices $K$ and $V$. 
\begin{equation}
Attention(Q,K,V)= softmax(\frac{QK^\top }{\sqrt{d_k}}) V
\end{equation}
\par 
where $\top$ is transpose function, $Q\in R^{n\times dk} , K \in R^{m\times dk}, V \in R^{m\times dv}$, $R$ is the real field, $n,m$ are the lengths of query and key/value sequences, $dk,dv$ are the dimensions of key and value. For summarization model we assume $K = V$. Self-attention can be defined from basic attention with $Q = K = V$. And multi-head attention concatenates multiple basic attentions with different parameters. We formulate multi-head attention as:

\begin{equation}
MH(Q,K,V) = Concat(hd_1,hd_2..., hd_i) w_{mh}
\end{equation}                        
where $hd_i = Attention(Qw_i^Q, Kw_i^K, Vw_i^V)$, and vector $w_i^Q, w_i^K, w_i^V, w_{mh}$ are learnable parameters.

\subsection{Transformer Baseline Model}
\label{ssec:transformer}
Our baseline model corresponds to the Transformer model in NMT tasks. The model is different from previous sequence-to-sequence models as it applies attention to replace RNN. The Transformer model can divide into encoder and decoder, and we will discuss them respectively below.\par 

\textbf{Input} The attention defined in the Transformer is the bag of words(BOW) model, so we have to add extra position information to the input. The position encodes with heuristic sine and cosine function:
\begin{align}
PE_{(pos,2i)} &= \sin(pos/10000^{2i/d_{model}})\\
PE_{(pos,2i+1)} &= \cos(pos/10000^{2i/d_{model}})
\end{align}
where $pos$ is the position of word in text, $i$ is the dimension index of embedding, and the dimension of model is $d_{model}$. The input of network $U$ is equal to source text word embeddings $E_w=\{e^1, e^2, ..., e^{ld}\}$ added position embeddings $E_p=\{p^1, p^2, ... p^{ld}\}$.\par 
 
\textbf{Encoder} The goal of encoder is extracting the features of input text and map it to a vector representation. The encoder stacks with $N$ encoder layer. Each layer consists of multi-head self-attention and position-wise feed-forward sublayers. We employ a residual connection around each of the two sublayers, followed by layer normalization. From the multi-head attention sublayer, we can extract different semantic information. Then we compute each encoder layer's final hidden states using position-wise feed-forward. The $l$th encoder layer is formulated as:\par 
\begin{equation}
\begin{aligned}
h_s^{(l)} &= Norm(MH(Q_s^{(l)}, K_s^{(l)}, V_s^{(l)}) + Q_s^{(l)})\\
h_{el}^{(l)} &= Norm(PFF(h_s^{(l)}) + h_s^{(l)}) \\
&= Norm((relu(h_s^{(l)}w_s^{l1}+b_s^{l1})w_s^{l2}+b_s^{l2}) + h_s^{(l)})
\end{aligned}
\end{equation}

where $h_s^{(l)}$ is the multi-head self-attention output after residual connection and $Norm(.)$ is layer normalization function, $h_{el}^{(l)}$ means the output of encoder layer ${l}$. $Q_s^{(l)}=K_s^{(l)}=V_s^{(l)}=U$ if $l=1$, or $Q_s^{(l)}=K_s^{(l)}=V_s^{(l)}=h_{el}^{(l-1)}$, vector $w_s^{l1}, w_s^{l2}$ and scalar $b_s^{l1}, b_s^{l2}$ are learnable parameters, and $PFF(.)$ is the position-wise feed-forward sublayer. This sublayer also can be described as two convolution operations with kernel size 1. \par

\textbf{Decoder} The decoder is used for generating salient and fluent text from the encoder hidden states. Decoder stacks with $N$ decoder layers. Each layer consists of masked multi-head self-attention, multi-head attention, and feed-forward sublayers. Similar to the encoder, we employ residual connections around each of the sublayers, followed by layer normalization. And we take $l$th decoder layer as example. We use the masked multi-head attention to encode summary as vector $h_{ms}^{(l)}$:

\begin{equation}
h_{ms}^{(l)} = Norm(MH^{*}(Q_{ms}^{(l)}, K_{ms}^{(l)}, V_{ms}^{(l)}) + h_{ms}^{(l)})
\end{equation}
where $Q_{ms}^{(l)}=K_{ms}^{(l)}=V_{ms}^{(l)}={(E_{gw}+E_{gp})}$ in the first layer and $Q_{ms}^{(l)}=K_{ms}^{(l)}=V_{ms}^{(l)}=h_{dl}^{(l-1)}$ in other layers. $h_{dl}^{(l-1)}$ is the output of the $(l-1)$th decoder layer, $E_{gw}, E_{gp}$ is the word embeddings and position embeddings of generated words respectively. The $MH^{*}(.)$ is masked multi-head self-attention and the mask is similar with the Transformer decoder. Then we execute multi-head attention between encoder and decoder:
\begin{equation}
h_d^{(l)} = Norm(MH(Q_d^{(l)}, K_d, V_d) + Q_d^{(l)})
\end{equation}
where $Q_d^{(l)}=h_{ms}^{(l)}$ is hidden states of decoder masked multi-head attention and $K_d=V_d=h_{el}^N$ is the last encoder layer output states. Finally, we use position-wise feed-forward and layer normalization sublayers to compute final states $h_{dl}^{(l)}$:
\begin{equation}
\begin{aligned}
h_{dl}^{(l)} &= Norm(PFF(h_d^{(l)}) + h_d^{(l)}) \\
&= Norm((relu(h_d^{(l)}w_d^{l1}+b_d^{l1})w_d^{l2}+b_d^{l2}) + h_d^{(l)})
\end{aligned}
\end{equation}
where vector $w_d^{l1}, w_d^{l2}$ and scalar $b_d^{l1}, b_d^{l2}$ are learnable parameters. And projecting the decoder final hidden states to vocab size then we can get vocabulary probability distribution $P_{vocab}$.\par

\subsection{Pointer and BPE Mechanism}
\label{ssec:oov}
In generation tasks, we should deal with the out of vocabulary(OOV) problem. If we do not tackle this problem, the generated text only contains a limited vocabulary words and replaces OOVs with $<unk>$. Things get worse in summarization task, the specific nouns(like name, place, etc.) with low frequency is the key information of summary, however, the vocabulary built with top $k$ words with the most frequent occurrence while those specific nouns may not occur in vocabulary. \par 
The pointer and byte pair encoder (BPE) mechanism are both used to tackle the OOV problem. The original BPE mechanism is a simple data compression technique that replaces the most frequent bytes pair with unused byte. Sennrich et al.\cite{sennrich2015neural} first use this technique for word segmentation via merging characters instead of bytes. So the fixed vocabulary can load more subwords to alleviate the problem of OOV.\par
The pointer mechanism allows both copying words from the source text and generating words from a fixed vocabulary. For pointer mechanism at each decoder time step, the generation probability $P_{gen} \in [0,1]$ can be calculated:
\begin{equation}
P_{gen} = \sigma(w_{dl}h_{dl}^N+b_{gen})
\label{eq: pgen}
\end{equation}
where vector $w_{dl}$ and scalar $b_{gen}$ are learnable parameter. $h_{dl}^N$ is the last decoder output states. We compute the final word distribution via pointer network: 
\begin{align}
\alpha &= softmax(h_{dl}^Nu^\top+b_{copy})\\
P_{copy} &= \sum_{1}^{ld}\alpha z_i\\
P_{final} &= P_{copy}(1-P_{gen}) + P_{vocab}P_{gen}
\end{align}
where $u$ is representation of input, $z_i$ is one-hot indicator vector for $w^i$, $P_{copy}$ is probability distribution of source words and $P_{final}$ is final probability distribution.\par 


\subsection{Aggregation Mechanism}
\label{ssec:aggregation}
The overview of our model is in Figure \ref{fig:base_model}. To enhance memory ability, we add the aggregation mechanism between encoder and decoder for collecting history information. The aggregation mechanism reconstructs the encoder's final hidden states by reviewing history information.  And we put forward two primitive aggregation approaches that can be proved effective in our task.\par

\begin{figure}
	\centerline{\includegraphics[width=\linewidth]{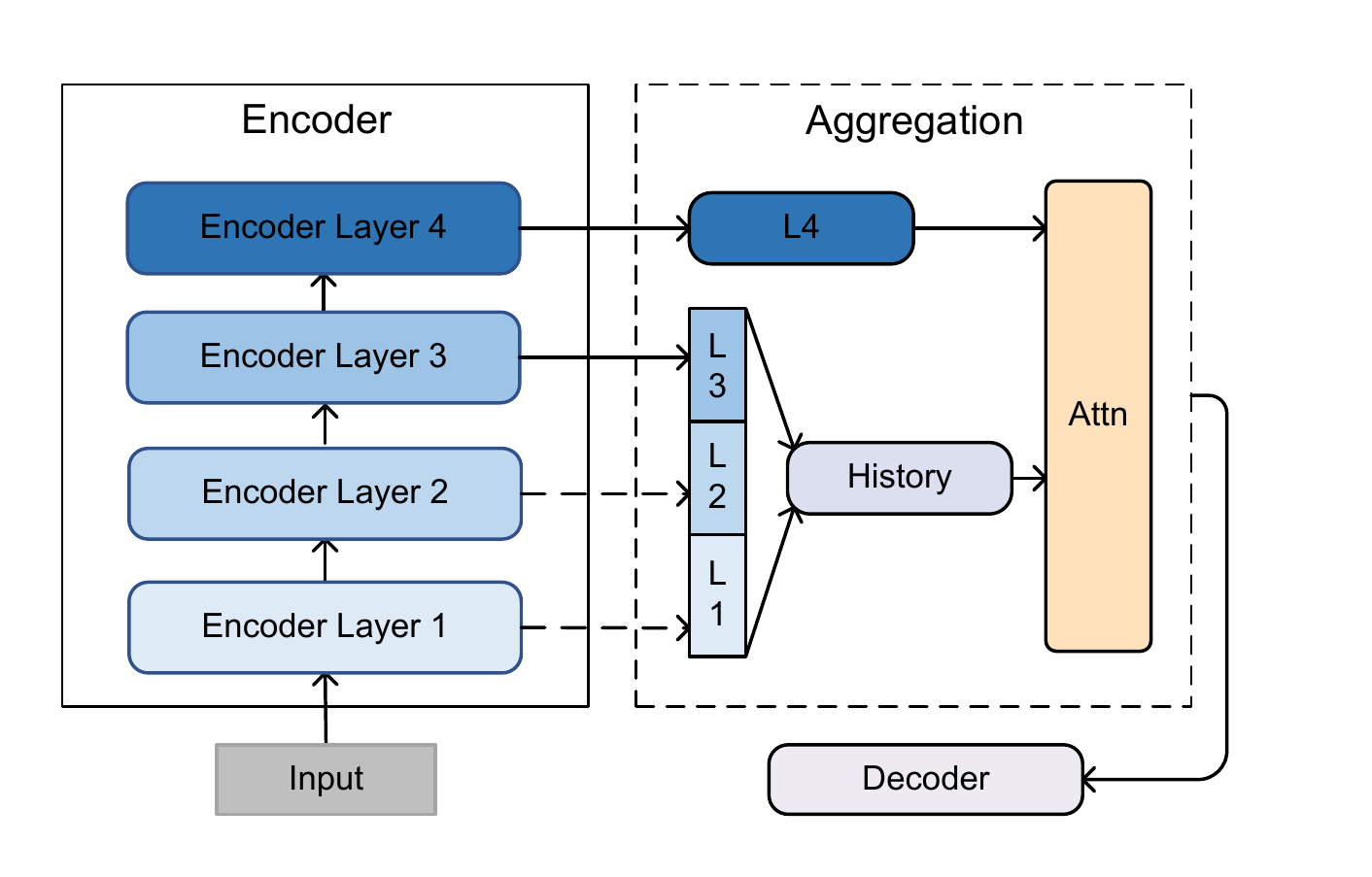}}
	\caption{The overview of projection aggregation mechanism with 4 encoder layers.}
	\label{fig:feed_forward}
\end{figure}

The first approach is using full-connected networks to collect historical information(see Figure\ref{fig:feed_forward}).
This approach first goes through normal encoder layers to get the outputs of each layer, and we select middle $L$ layers' outputs then concatenate them as input of full connected networks to obtain history information $H = h^h$. Finally, we compute multi-head attention between history state $H$ and the output of the last encoder layer.
This process can be formulated as:
\begin{equation}
h^h = w^h(Concat(h_{el}^{(N-L)},...,h_{el}^{(N-1)})) + b^h
\end{equation} 
where vector $w^h$ and scalar $b^h$ are learnable parameters, $L$ is hyper-parameter to be explored.  Then we add the multi-head attention layer between the last encoder layer output $h_{el}^N$ and history information $h^h$. The output of attention is the final states of encoder:
\begin{equation}
h^a = MH(Q^p, K^p, V^p) 
\end{equation}
where $Q^p$ is history information $h^p$ and $K^p=V^p=h_{el}^N$. 

\begin{figure}
	\centerline{\includegraphics[width=\linewidth]{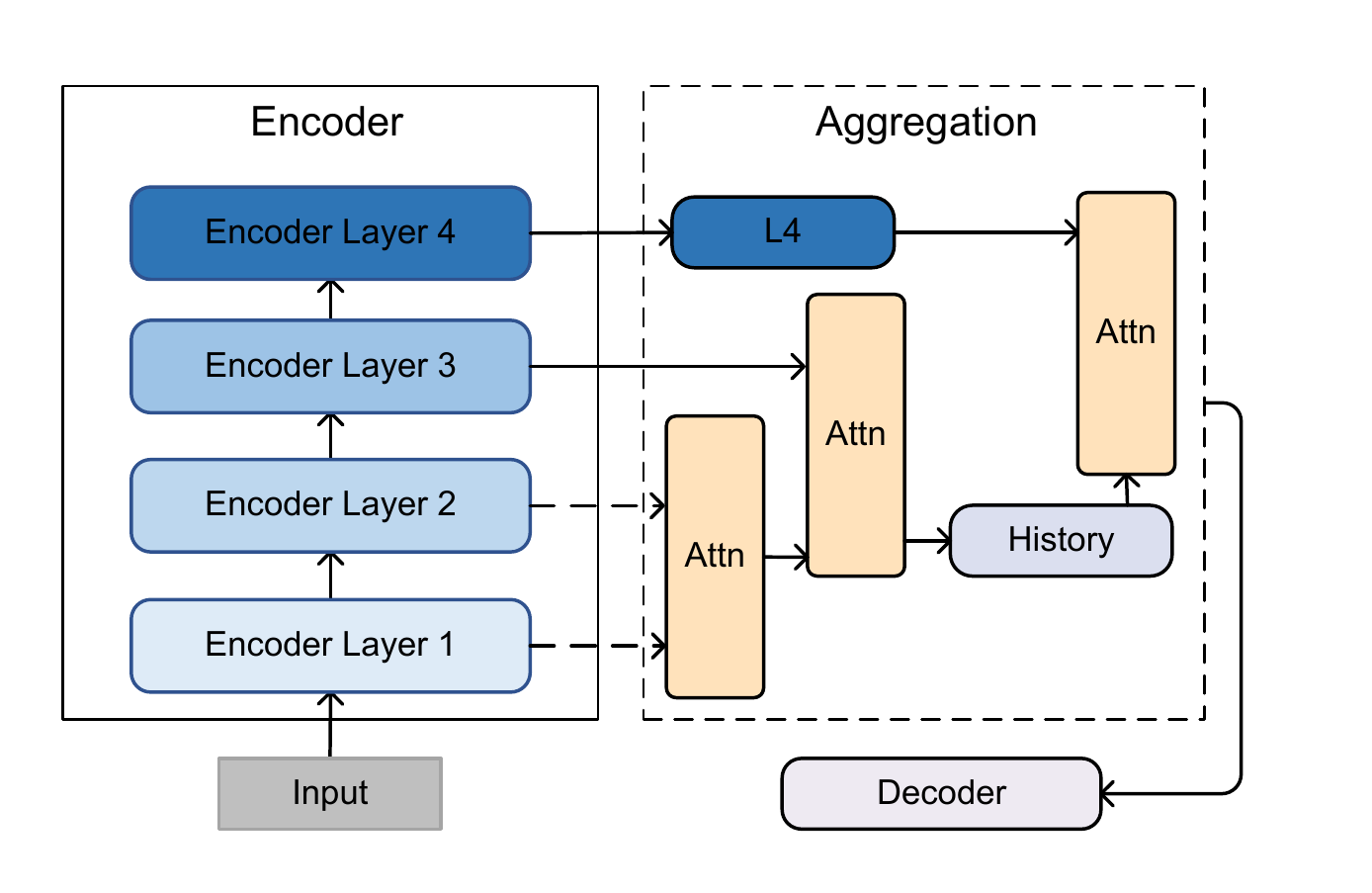}}
	\caption{The overview of attention aggregation mechanism with 4 encoder layers.}
	\label{fig:attention}
\end{figure}

The second approach is using attention mechanism to collect history information(see Figure \ref{fig:attention}). We select middle $L$ encoder layers' outputs to iteratively compute multi-head attention between current encoder layer output and previous history information. And the $l$th history information $h^{h(l)}$ can be calculated as follows:
\begin{equation}
h^{h(l)} = MH(Q^{p(l)}, K^{p(l)}, V^{p(l)}) 
\end{equation}
where $l \in [N-L, N)$ is index of selected encoder layers, $Q^{p(l)}$ is previous history state $h^{h(l-1)}$ and $K^{p(l)} = V^{p(l)}$ is encoder output $h_{el}^l$ . 
Iteratively calculating history information until the last selected encoder layer, we can get final history hidden states $h^a$ and make the states as the final states of the encoder.\par 
Finally, we define the objective function. Given the golden summary $Y={y_1,y_2,...y_{ly}}$ and input text $X$, we minimize the negative log-likelihood of the target word sequence. The training objective function can be described:
\begin{equation}
J(\theta) = \sum_{k=1}^{ly} - \log p(Y|X;\theta)
\end{equation}
where $\theta$ is model parameter and $N$ is the number of source-summary text pairs in training set. 
The loss for one sample can be added by the loss of generated word $y_t$ in each time step $t$:
\begin{equation}
\log(Y|X;\theta) = \sum_{t=1}^{T}\log p(y_t|y_1,y_2,...y_{(t-1)},X;\theta)
\end{equation}
where $p(y_t|y_1,y_2,...y_{(t-1)},X;\theta)$ can be calculated in decoder $t$ time step, $T$ is total decoding steps. 

\section{EXPERIMENTS}
\label{sec:experiments}
In this section, we first define the setup of our experiment and then analyze the results of our experiments.
\subsection{Experimental Setup}
\label{subsec:experiments_setup}
\quad\textbf{Dataset}
We conduct our experiments on CNN/DailyMail dataset\cite{hermann2015teaching, nallapati2016abstractive}, which has been widely used for long document summarization tasks. The corpus is constructed by collecting online news articles and human-generated summaries on CNN/Daily Mail website. We choose the non-anonymized version\footnote[1]{\url{https://github.com/abisee/cnn-dailymail}}\cite{see2017get}, which is not replacing named entity with a unique identifier. The dataset contains pairs of articles and summaries. The details of this dataset are in section \ref{subsec: result}.\par 
\begin{table*}
	\begin{center}
		\caption{Comparison of different model results on CNN/DaliyMail test dataset using F1 scores of ROUGE-1, ROUGE-2, ROUGE-L with $ 95\%$ confidence interval. The first part is previous abstractive baseline models, the second part is the Transformer baseline model and our Transformer model with aggregation mechanism. The best scores are bolded.}
		\label{table:result}
		\begin{tabular}{|l|c|c|c|}
			\hline
			Model & ROUGE-1 & ROUGE-2 & ROUGE-L  \\
			\hline
			lead-3 & 40.24 & 17.52 & 36.34 \\
			words-1vt2k-temp-att \cite{nallapati2016abstractive} & 36.64 & 15.66 & 33.42\\
			ConvS2S \cite{gehring2017convolutional} & 39.75 & 17.29 & 36.54 \\
			Pointer Generator + Coverage \cite{see2017get}& 39.53 & 17.28 & 36.38 \\
			Pointer Generator + Coverage + cbdec + RL\cite{jiang2018closed-book} & 40.66 & 17.87 & 37.06 \\
			Inconsistency Loss \cite{hsu2018a} & 40.68 & 17.97 & 37.13 \\
			rnn-ext + abs + RL + rerank \cite{chen2018fast} & 40.88 & 17.80 & \textbf{38.54}\\
			\hline
			Transformer & 40.05 & 17.72 & 36.77 \\
			
			
			Aggregation Transformer(attention) & \textbf{41.06} & \textbf{18.02} & \textbf{38.04} \\
			
			\hline
		\end{tabular}	
		
	\end{center}
\end{table*}

\textbf{Training Details}
We conduct our experiments with 1 NVIDIA Tesla V100. During training and testing time we truncate the source text to $500$ words and we build a shared vocabulary for encoder and decoder with small vocabulary size $50$k, due to the using of the pointer or BPE mechanism. Word embeddings are learned during training time. We use Adam optimizer with initial learning rate $10^{-4}$ and parameter $\beta_1=0.9, \beta_2=0.999$ in training phase. We adapt the learning rate according to the loss on the validation set (half learning rate if validation set loss is not going down in every two epochs). And we use regulation with all $dropout = 0.1$. The training process converges about 200,000 steps for each model.\par
In the generation phase, we use the beam search algorithm to produce multiple summary candidates in parallel to get better summaries and add repeated words to blacklist in the processing of search to avoid duplication. For fear of favoring shorter generated summaries, we utilize the length penalty. In detail, we set beam size $10$, no-repeated n-gram size  $3$ and length penalty parameter $2.0$. We also constrain the maximum and minimum length of the generated summary to $120$ and $50$ respectively.\par 

We evaluate our system using F-measures of ROUGE-1, ROUGE-2, ROUGE-L metrics which respectively represent the overlap of N-gram and the longest common sequence between the golden summary and the system summary. The scores are computed by python pyrouge\footnote[2]{\url{https://pypi.org/project/pyrouge/}} package.\par

\textbf{Experiment explorations}
We explore the influence of different experiment hyper-parameters setup for the model's performance, which includes 11 different experiment settings. 

Firstly, we explore the number of Transformer encoder/decoder layers (see Table \ref{table:layers}).

Secondly, we dig out the different aggregation methods with 1 aggregation layer 
(see Table \ref{table:ablation_agg}). The exploration includes our baseline model(\textbf{m1}) and Transformer model with add function(\textbf{m2}), projection aggregation method(\textbf{m4}) and attention aggregation method(\textbf{m6}). \par
Thirdly, we also explore the different performance of different number of aggregation layers (see Table \ref{table:ablation_agg}). There are 3 groups of experiments with different number of aggregation layers: Transformer adding last 2 layers(\textbf{m2}) and last 3 layers(\textbf{m3}), Transformer with projection aggregation method using 1 layer(\textbf{m4}) and 2 layers(\textbf{m5}) and Transformer with attention aggregation method using 1 layer(\textbf{m6}) and 2 layer(\textbf{m7}). 
For all models except the exploration of encoder/decoder layers, we use 4 encoder and 4 decoder layers.

\textbf{Human Evaluation} 
The ROUGE scores are widely used in the automatic evaluation of summarization, but it has great limitations in semantic and syntax information. In this case, we use manual evaluation to ensure the performance of our models. We perform a small scale human evaluations where we randomly select about 100 generated summaries from each of the 3 models(Pointer Generator, Transformer, and aggregation Transformer) and randomly shuffle the order of 3 summaries to anonymize model identities, then let 20 anonymous volunteers with excellent English literacy skills score random 10 summaries for each 3 models range from 1 to 5(high score means high-quality summary). then we using the average score of each summary as their final score.   
the evaluation criteria are as follows: (1) salient: summaries have the important point of the source text, (2) fluency: summaries are consistent with human reading habits and have few grammatical errors, (3) non-repeated: summaries do not contain too much redundancy word.

\subsection{Results}
\label{subsec: result}

\begin{table}
	\begin{center}
		\caption{The comparison of translation and summarization datasets. We remove sentence tags in the source text and split sentences with blank, then count maximal and average length token in each dataset.}
		\label{table: dataset}
		\resizebox{\linewidth}{!}{
		\begin{tabular}{|c|c|c|c|}
			\hline
			Dataset & Train & Valid & Test \\
			\hline
			CNN/DailyMail(summarization) & 287226 & 13368 & 11490 \\
			max-token-len(art/abs) & 2882 / 2096 & 2134 / 1684 & 2377 / 678 \\
			avg-token-len(art/abs) & 790 / 55 & 768 / 61 & 777 / 58 \\
			\hline
			Our Dataset(summarization) & 48600 & 4800 & 6600 \\
			max-token-len(art/abs) & 1914 / 80 & 1687 / 80 & 1670 / 80\\
			avg-token-len(art/abs) & 768 / 65 & 763 / 65 & 769 / 65\\
			\hline
			iwslt14-de-en(translation) & 160239 & 7283 & 6750 \\
			max-token-len(de/en) & 244 / 228 & 169 / 154 & 245 / 217 \\
			avg-token-len(de/en) & 24 /24 & 24 / 24 & 23 / 22 \\
			\hline
			wmt17-en-de(translation) & 3961179 & 40058 & 3003 \\
			max-token-len(en/de) & 250 /250 & 224 / 233 & 101 / 93 \\
			avg-token-len(en/de) & 28 /29 & 28 / 29 & 26 / 27 \\
			\hline
		\end{tabular}}
		
	\end{center}	
\end{table}

\quad\textbf{Dataset Analysis} 
To demonstrate the difference between summarization and translation tasks, we compare the dataset for two tasks (see Table \ref{table: dataset}). The summarization dataset CNN/DailyMail contains 287226 training pairs, 13368 validation pairs, and 11490 test pairs. The translation dataset iwslt14 and wmt17 have 160239/3961179 training pairs, 7283/40058 validation pairs, and 6750/3003 test pairs respectively. Then we find the characteristics of those two different tasks after comparison. The summarization source text can include more than 2000 words and the average length of the source text is 10 times longer than the target text, while the translation task contains at most 250 words and the average length of the source text is about the same as the target text. Because of that, we need a strong encoder with memory ability to decide where to attend and where to ignore. \par 
 \begin{figure}[ht]
	\begin{center}
		\begin{tabular}{|p{0.9\columnwidth}|}
			\hline
			\textbf{Source Text(truncated 500)}:
			(......) \textcolor{magenta}{national grid has revealed the uk 's first new pylon for nearly 90 years .} called the t-pylon \textcolor{cyan}{-lrb- artist 's illustration shown -rrb-} it is a third shorter than the old lattice pylons . \textcolor{green}{but it is able to carry just as much power - 400,000 volts . it is designed to be less obtrusive and will be used for clean energy purposes .} national grid is building a training line of the less obtrusive t-pylons at their eakring training academy in nottinghamshire . britain 's first pylon , erected in july 1928 near edinburgh , was designed by architectural luminary sir reginald blomfield , inspired by the greek root of the word ` pylon ' -lrb- meaning gateway of an egyptian temple -rrb- . the campaign against them - they were unloved even then - was run by rudyard kipling , john maynard keynes and hilaire belloc . five years later , the biggest peacetime construction project seen in britain , the connection of 122 power stations by 4,000 miles of cable , was completed . 
			it marked the birth of the national grid and was a major stoking of the nation 's industrial engine and a vital asset during the second world war (......) \\
			\hline
			\textbf{Ground Truth:}
			national grid has revealed the uk 's first new pylon for nearly 90 years .
			called the t-pylon it is a third shorter than the old lattice pylons .
			but it is able to carry just as much power - 400,000 volts .
			it is designed to be less obtrusive and will be used for clean energy .\\
			\hline
			\textbf{Transformer Baseline:}
			the t-pylon -lrb- artist 's shown -rrb- it is a third shorter than the old lattice pylons .
			but it is able to carry just as much power - 400,000 volts .
			it is designed to be less obtrusive and will be used for clean energy purposes .\\
			\hline
			\textbf{Our model:}
			national grid has revealed the uk 's first new pylon for nearly 90 years .
			called the t-pylon it is a third shorter than the old lattice pylons .
			but it is able to carry just as much power - 400,000 volts .
			it is designed to be less obtrusive and will be used for clean energy purposes .\\
			\hline
			
		\end{tabular}	
		\caption{The comparison of ground truth summary and generated summaries of 2 abstractive summarization models on CNN/DailyMail dataset. The red represents missed information, the blue means unnecessary information and the green signify appropriate information. }
		\label{fig:result}
	\end{center}
\end{figure}

\begin{figure}
	
	\centerline{\includegraphics[width=\columnwidth]{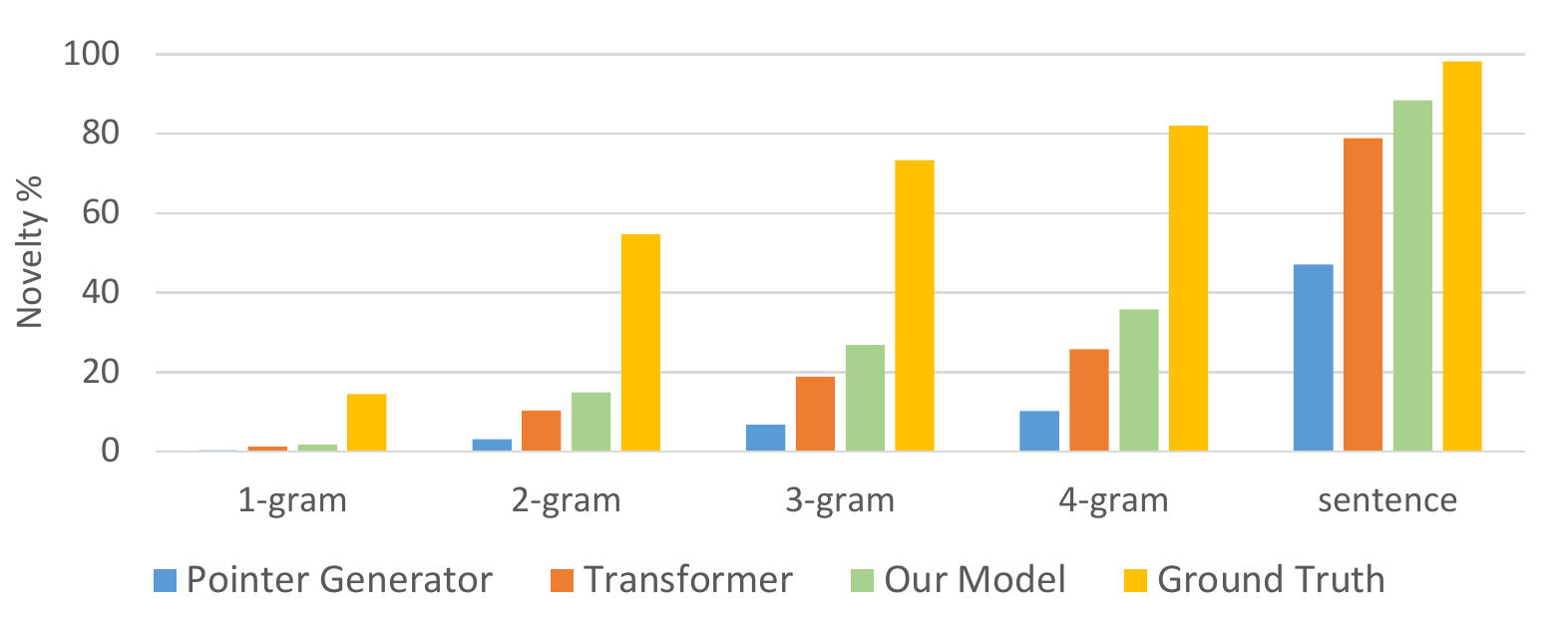}}
	\caption{The statistics of novel n-grams and sentences. Our model can generate far more novel n-grams and sentences than Pointer Generator and Transformer baseline.}
	\label{fig:novetly}
\end{figure}

\textbf{Quantitative Analysis} The experimental results are given in Table \ref{table:result}. Overall, our model improves all other baselines(reported in their articles) for ROUGE-1, 2 F1 scores, while our model gets a lower ROUGE-L F1 score than the RL (Reinforcement Learning) model\cite{chen2018fast}. From celikyilmaz et al.\cite{celikyilmaz2018deep}, the ROUGE-L F1 score is not correlated with summary quality, and our model generates the most novel words compared with other baselines in novelty experiment \ref{fig:novetly}. The novel words are harmful to ROUGE-2, L F1 scores. This result also account for our models being more abstractive.\par  
Figure \ref{fig:result} shows the ground truth summary, the generated summaries from the Transformer baseline model and our aggregation Transformer using the attention aggregation method. The source text is the main fragment of the truncated text. Compared with the aggregation Transformer, the summary generated by the Transformer baseline model have two problems. Firstly, the summary of the baseline model is lack of salient information marked with red in the source text. Secondly, it contains unnecessary information marked with blue in the source text. \par 
we hold the opinion that the Transformer baseline model has weak memory ability compared to our model. Therefore, it can not remind the information far from its current states which will lead to missing some salient information and it may remember irrelevant information which will lead to unnecessary words generated in summaries. Our model uses the aggregation mechanism that can review the primitive information to enhance the model memory capacity. Therefore, the aggregation mechanism makes our model generate salient and non-repetitive words in summaries.\par
 
\begin{table}
	\begin{center}
		\caption{We compare different layers of encoder(E) and decoder(D) and report results on CNN/DailyMail test dataset using precision/recall/F1 scores of ROUGE.}
		\label{table:layers}
		\resizebox{\linewidth}{!}{
			\begin{tabular}{|c|ccc|ccc|ccc|}
				\hline
				E/D & \multicolumn{3}{c|}{ROUGE-1(P/R/F1)} & \multicolumn{3}{c|}{ROUGE-2(P/R/F1)} & \multicolumn{3}{c|}{ROUGE-l(P/R/F1)} \\
				\hline
				4/4 &40.46&41.53&40.05&18.11&18.42&17.72&36.42&37.15&36.77 \\
				4/3 &40.88&40.47&39.75&18.40&17.93&17.63&37.07&36.70&36.50 \\
				4/2 &41.70&39.23&39.54&18.78&17.26&17.47&37.96&35.87&36.51 \\
				2/4 &39.88&41.26&39.57&17.67&18.07&17.30&35.97&37.00&35.98 \\
				3/4 &40.46&40.01&39.80&18.05&18.10&17.54&36.63&37.07&36.43 \\
				\hline
		\end{tabular}
	}
		
	\end{center}
\end{table}

\textbf{Encoder/Decoder Layers Analysis}  
The first exploration experiment consists of Transformer models using different encoder and decoder layers. And we only experiment if the number of encoder/decoder layers is no more than 4.
We also tried 6 encoder and decoder layers, however, there is no notable difference with 4 encoder and decoder layers and increasing a lot of parameters and taking more time to converge. Therefore we make the Transformer baseline model have 4 encoder and decoder layers. \par  
If we decrease the layers of encoder or decoder respectively, the results are shown in Table \ref{table:layers}. 
It can be concluded from the comparison of each model results that we can get lower precision but higher recall score when the encoder layers are decreasing and we have opposite results on the decoder layers decreasing experiments. 
Meanwhile, we can get a higher ROUGE-1 F1 score and lower ROUGE-2, L F1 scores in the model decreasing each 1 decoder layer compared to that decreasing each 1 encoder layer. 
Therefore, we can conclude that the encoder captures the features of the source text while the decoder makes summaries consistently. \par 
 
\begin{table}
	\caption{The aggregation mechanism experiments. our experiments use 3 aggregation methods with 2 different aggregation layers.}
	\label{table:ablation_agg}
	\resizebox{\linewidth}{!}{
		\begin{tabular}{|l|c|c|c|}
			\hline
			Model & ROUGE-1 & ROUGE-2 & ROUGE-L  \\
			\hline
			(m1)Transformer & 40.05 & 17.72 & 36.77 \\
			(m2)Transformer(add 1 layer) & 39.79 & 17.52 & 36.32 \\
			(m3)Transformer(add 2 layer) & 39.69 & 17.34 & 36.15 \\
			(m4)Agg-Transformer(proj 1 layer) & 40.58 & 17.77 & 36.60 \\
			(m5)Agg-Transformer(proj 2 layer) & 40.67 & 17.84 & 36.70 \\
			(m6)Agg-Transformer(attn 1 layer) & \textbf{41.06} & \textbf{18.02} & \textbf{38.04} \\
			(m7)Agg-Transformer(attn 2 layer) & 40.03 & 17.59 & 36.60 \\
			\hline
		\end{tabular}
	}
	
\end{table}

\textbf{Aggregation mechanism Analysis} 
The second exploration experiment consists of our baseline model(\textbf{m1, m2}) and aggregation Transformer model using different aggregation mechanism(\textbf{m4, m6}) in Table \ref{table:ablation_agg}. If we use baseline model adding the last $L$ layer(s) simply(\textbf{m2}), the result scores will decrease beyond our expectation.
However, simply adding the last $L $ layer(s) can re-distribute the encoder final states with history states, it will average the importance weights of those layers and that maybe get things worse. 
Compared with the baseline model, the result scores of our aggregation models(\textbf{m4, m6}) are boosting. We compute attention between history(query) and encoder final states(key/value) to re-distribute the final states so that the encoder obtains the ability to fusing history information with different importance.

The third exploration contains 3 groups experiments: add group(\textbf{m2, m3}), projection group(\textbf{m4, m5}) and attention group(\textbf{m6, m7}). The aggregation Transformer models here use different aggregation layers. We also experiment with the model in the above 3 groups with 3 aggregation layers, but they all get extraordinary low ROUGE scores (all 3 models have ROUGE-1 39.3, ROUGE-2 14.5, ROUGE-L 34.3 roughly). They all incorporate the output of the first encoder layer which may not have semantic information which may be harmful to the re-distributing of the encoder final states. So we do not compare with those models explicitly.\par 

For add aggregation group, we increase the added layers while the ROUGE scores will get down. If we add more layers, the final state distributions will tend to be the uniform distribution which makes decoder confused about the key ideas of source text. For that reason, we may get worse scores when we add more layers.\par 

For the projection aggregation group, we increase the aggregation layers and the ROUGE scores will rise. If we aggregate more layers, the history states will contain more information which will lead to performance improvement. However,
we will lose a lot of information when the aggregation layers increasing. And we achieve the best result with 2 aggregation layers.\par 

For the attention aggregation group, we get the best score with 1 aggregation layer but the ROUGE scores will decline if we increase the aggregation layers. We just need one layer attention to focus on history states, because too much attention layers may have an excessive dependency on history states. If the encoder final distribution focus more on shallow layers which introduced a lot of useless information, it is harmful to the encoder to capture salient features.\par


\textbf{Abstractive analysis} 
Figure \ref{fig:novetly} shows that our model copy $10\%$ whole sentences from source texts, and the copy rate is almost close to reference summaries. However, there is still a huge gap in n-grams generation, and this is the main area for improvement. \par 
In particular, the Pointer Generator model tends to examples with few novel words in summaries because of its lower rate of novel words generation. The Transformer baseline model can generate novel summaries and our model get great improvement (with 0.5, 4.6, 7.8, 10.1\% novelty improvement for n-gram($n\in\{1,2,3,4\}$)) compared to the Transformer baseline model. Because our model reviews history states and re-distribute encoder final states, we get more accurate semantic representation. It also proves that our aggregation mechanism can improve the memory capability of encoder.\par
 
\begin{table}
	\begin{center}
		\caption{Human evaluation of three models. We compare the average score of salient, fluency and non-repeated. The best scores are bolded.}
		\label{table:human_eva}
		\begin{tabular}{|l|c|c|c|}
			\hline
			Model & Salient & Fluency & Non-Repeated  \\
			\hline
			Pointer Generator & 3.37 & 3.12 & 3.17\\
			Pointer Generator + Coverage& 3.42 & 3.23 & 3.61\\
			Transformer & 3.56 & 3.30 & 3.67 \\
			Transformer + Aggregation & \textbf{3.87} & \textbf{3.37} & \textbf{3.78}\\
			\hline
		\end{tabular}
	\end{center}

\end{table}

\textbf{Human Evaluation} 
We conduct our human evaluation with setup in section \ref{subsec:experiments_setup}, and the results show in Table \ref{table:human_eva}. We only compared three models on salient, fluency and non-repeated criteria, and our model gets the highest score in all criteria. But in fluency criterion, none of the models scores well, which means it is hard to understand semantic information for all models now. 
The Pointer Generator is our baseline abstractive summarization approach and has the lowest scores. The Pointer Generator uses the coverage mechanism to avoid generating overlap words, which can make summaries more fluent and less repetitive. The transformer is a new abstractive summarization based on attention mechanism, and it can get better performance than the Pointer Generator model. We equip the Transformer model with the aggregation mechanism, and it can get great improvement on all 3 criteria. 

\subsection{Our Chinese Experiments}
We build our Chinese summarization dataset via crawling news website\footnote{\url{https://www.thepaper.cn/}} and process the raw web page contents to character-based texts. The details of our dataset show in Table \ref{table: dataset} where our dataset has a similar average length of source texts and summaries compared CNN/DM dataset. It is a temporary dataset, which only contains 60,000 pairs of text totally for now, and we are still adding data to our dataset. \par 
\begin{table}
	
	\begin{center}
		\caption{Experiments on our Chinese dataset.
			We only experiment on three baseline models and evaluate results with ROUGE F metrics. The best scores are bolded.}
		\label{table:chinese_result}
			\begin{tabular}{|l|c|c|c|}
				\hline
				Model & ROUGE-1 & ROUGE-2 & ROUGE-L  \\
				\hline
				Lead-3 & 54.09 & 42.46 & 34.56\\
				Pointer Generator & 55.49 & 43.59 & 48.03\\
				Pointer Generator + Coverage& 55.64 & 43.80 & 48.08\\
				Transformer & 52.69 & 39.86 & 43.66 \\
				Transformer + Aggregation & \textbf{58.00} & \textbf{44.42} & \textbf{48.85} \\
				\hline
			\end{tabular}	
	\end{center}

\end{table}
We also experiment on our Chinese dataset and evaluate the result with ROUGE\footnote{\url{https://github.com/tensorflow/tensor2tensor/blob/master/tensor2tensor/utils/rouge.py} 
} metrics. Our model gets the highest score, while the Pointer Generator model gets rather high ROUGE scores (see Table \ref{table:chinese_result}). Because the dataset does not contain many novel words where it is suitable for the Pointer Generator model. Our dataset contains (6.17, 14.51, 17.99, 20.10)\% novel (1,2,3,4)-gram and 59.90\% novel sentences; by comparison, the novel n-gram and sentences frequency of CNN/DM in Figure \ref{fig:novetly} is (14.47, 54.75, 73.32, 82, 98.16)\% respectively. And the Pointer Generator model generates summaries containing less novel words and sentences, which leads to high scores in our Chinese dataset. Finally, we compare our model with the Transformer baseline model, and our results improve 5.31 in ROUGE-1, 4.56 in ROUGE-2 and 5.19 in ROUGE-L scores.

\section{CONCLUSIONS}
\label{sec:conclusions}
In this paper, we propose a new aggregation mechanism for the Transformer model, which can enhance encoder memory ability. The addition of the aggregation mechanism obtains the best performance compared to the Transformer baseline model and Pointer Generator on CNN/DailyMail dataset in terms of ROUGE scores. 
We explore different aggregation methods: add, projection and attention methods, in which attention method performs best. We also explore the performance of different aggregation layers to improve the best score. We build a Chinese dataset for the summarization task and give the statistics of it in Table \ref{table: dataset}. our proposed method also achieves the best performance on our Chinese dataset.\par  
In the future, we will explore memory network to collect history information and try to directly send history information to the decoding processing to improve the performance in the summarization task. And the aggregation mechanism can be transferred to other generation tasks as well. 

\section*{ACKNOWLEDGMENT}
\label{sec:ack}
This work was supported by the National Natural Science Foundation of China (Grant No.61602474).
\clearpage
\bibliography{ecai}
\end{document}